\documentclass[10pt,conference]{IEEEtran}

\usepackage{amsmath,amssymb,amsfonts,amsthm}
\usepackage{booktabs}
\usepackage{mathtools}
\usepackage[hidelinks]{hyperref}
\usepackage[T1]{fontenc}
\usepackage{newtxtext,newtxmath}
\usepackage{xcolor}

\usepackage{hyperref}
\usepackage{caption}
\usepackage{balance}
\usepackage{stfloats}

\usepackage{float}

\definecolor{redC}{RGB}{190,40,35}
\definecolor{greenC}{RGB}{40,150,70}
\definecolor{blueC}{RGB}{40,80,190}

\title{CPrefix: A Combinatorial Tensor Framework for Structured Discrete Color Mappings}

\author{%
\IEEEauthorblockN{Yvan Richard}
\IEEEauthorblockA{
\href{https://orcid.org/0000-0003-4497-3843}
{ORCID: 0000-0003-4497-3843}
\vspace{-2.0em}
}
}

\begin{document}

\maketitle
\footnotetext{
An earlier version of this work was peer-reviewed and accepted for presentation at the IEEE ICIP 2026 Workshop on Computational Color Imaging (CCIW 2026). It was withdrawn before publication in the conference proceedings because the author was unable to attend. This arXiv manuscript incorporates revisions based on the reviewers' comments.
}

\begin{abstract}
Discrete multi-channel mappings are typically represented through sampled values, providing accurate evaluations but limited insight into their underlying structure. We introduce CPrefix, a combinatorial observable representation for discrete mappings, realized within a unified tensor framework that enables representation, reconstruction, and structural analysis.

\vspace{0.25em}
\noindent
The framework is based on a counting tensor induced by multinomial counting observables. Its support forms a discrete Pascal simplex, not as a constraint on the observable space, but as a latent combinatorial representation from which mappings are reconstructed. This formulation separates the combinatorial organization of a mapping from its measured values, exposing the observable structure underlying the mapping.

\vspace{0.25em}
\noindent
The framework is validated on ICC display and printer profiles through latent reconstruction and perceptual gamut transport. Accurate reconstruction demonstrates that color mappings admit faithful observable representations, while reconstruction residuals provide insight into the compatibility of the underlying mapping with the proposed representation.

\vspace{0.25em}
\noindent
Although demonstrated on color transformations, the framework is independent of the physical interpretation of the observables, making it applicable to structured multi-channel mappings arising from color imaging, spectral measurements and other discrete systems.
\end{abstract}

\vspace{-0.2em}
\section{Introduction}

Discrete multi-channel mappings arise throughout science and engineering whenever observable configurations are related by measured or computed transformations. Color imaging provides a representative example through lookup tables, interpolation schemes, and profile-based transforms relating input and output color spaces. While these representations accurately evaluate mappings, they operate at the level of sampled values and provide limited insight into the observable structure underlying the mapping itself.

\vspace{0.3em}
\noindent
We introduce \emph{CPrefix}, a combinatorial observable representation for structured discrete mappings. Rather than treating a mapping as a collection of sampled values, \emph{CPrefix} represents its observable organization within a unified tensor framework, simultaneously providing representation, reconstruction, and a quantitative framework for analyzing its underlying combinatorial structure. This perspective shifts the emphasis from evaluating discrete mappings to understanding their observable organization.

\vspace{0.25em}
\noindent
The framework is realized through a counting tensor induced by multinomial counting observables. The observable domain remains unchanged throughout the construction. Its support forms a discrete Pascal simplex defined by the admissibility condition $r+g+b \le N$, which should not be interpreted as a physical constraint imposed on the observable space, but as a latent combinatorial coordinate system organizing observable configurations according to their counting structure. 

\vspace{0.25em}
\noindent
Observable configurations remain defined in the original measurement space; the simplex provides an alternative combinatorial coordinate system for representing them. The tensor maps these simplex configurations to observable states. Rather than representing only measured values, the construction represents the observable structure underlying the mapping itself.

\vspace{0.25em}
\noindent
Concretely, for each observable configuration $(R,G,B)$ and simplex configuration $(r,g,b)$, the tensor counts the number of compatible ternary sequences $\Phi=\{0,1,2\}^{N}$. 

\vspace{0.25em}
\noindent
Equivalently, it aggregates lattice paths on the discrete Pascal simplex, generalizing the Hamming-weight shell decomposition of the binary hypercube to a multinomial setting, where observable shells correspond to combinatorial classes of equivalent ternary paths~\cite{MacWilliamsSloane,feller}.

\vspace{0.35em}
\noindent
\textit{We lift the discrete Pascal-simplex structure of admissible multinomial configurations into an observable multi-channel domain through a tensor mapping.}

\vspace{0.35em}
\noindent
\textbf{Figure~\ref{fig:CQDICE}} illustrates the relationship between discrete trajectory ensembles, Pascal-simplex multiplicities, and observable configurations. The framework separates intrinsic combinatorial organization from device-dependent behavior: the simplex provides the latent combinatorial representation, while the tensor describes how this representation is projected into measured space. Reconstruction residuals therefore reflect not only reconstruction fidelity but also the compatibility of the underlying mapping with the proposed observable representation.

\vspace{0.3em}
\noindent
Previous work~\cite{Richard2024} introduced a Pascal-based cellular automaton derived from pairwise observations. The present work shows that it corresponds to a degenerate contraction of a more general combinatorial tensor in which the admissible configuration space collapses to a one-dimensional marginal. Here, the formulation is extended to a general observable representation for structured discrete mappings.

\vspace{0.4em}
\noindent
\textit{Although demonstrated on three-channel color mappings, the framework is independent of the physical interpretation of the observables and naturally extends to arbitrary multi-channel discrete mappings, including color imaging, spectral measurements, and other structured observable systems.}

\clearpage
\section{CPrefix Framework}

\subsection{Notation}
\begin{table}[!h]
\centering
\small
\vspace{-0.8em}
\scriptsize
\setlength{\tabcolsep}{3pt}

\begin{tabular}{ll}
\toprule
Symbol & Meaning \\
\midrule

$\Phi$ & Ternary configuration ensemble $\{0,\dots,d-1\}^N$ \\
$\phi$ & Individual discrete path/configuration in $\Phi$ \\
$x(\phi)$ & Count projection into $\mathcal{C}_N^{(d)}$ \\

$\mathcal{C}_N^{(d)}$ & Observable domain (discrete cube) \\
$\mathcal{I}_N^{(d)}$ & Inner/core domain (discrete simplex) \\
$\mathcal{O}_N^{(d)}$ & Outer observable domain \\

$M$ & $|\mathcal{C}_N^{(d)}|=(N+1)^d$ \\
$K$ & $|\mathcal{I}_N^{(d)}|=\binom{N+d}{d}$ \\
$Q$ & $|\mathcal{O}_N^{(d)}|=M-K$ \\

$T$ & Structured counting tensor/operator \\
$T_{\mathrm{II}}$ & Inner-to-inner tensor block $\in\mathbb{R}^{K\times K}$ \\
$T_{\mathrm{OI}}$ & Outer-to-inner tensor block $\in\mathbb{R}^{Q\times K}$ \\

$C_{\mathrm{inner}}$ & Inner/simplex coordinates $\in\mathbb{R}^{K}$ \\
$X_{\mathrm{obs}}$ & Observable representation $\in\mathbb{R}^{M}$ \\

$\pi_{\mathrm{shell}}$ & Shell permutation ordering indices by $\ell_1$ shell level \\

$S$ & Observable shell coordinate $S=R+G+B$ \\
$s$ & Inner/simplex shell coordinate $s=r+g+b$ \\

$\mathrm{T}(R,G,B,r,g,b)$
& Full observable tensor \\

$T(R,G,B,s)$
& Diagonally contracted tensor \\

$T(S,s)$
& Shell-contracted tensor \\

$K_{\mathrm{Trgb}}$
& Induced observable kernel
$\mathrm{T}^{\top}\mathrm{T}$ \\

$C_{\mathrm{inner}}$
& Reconstructed latent simplex coordinates \\
\bottomrule
\end{tabular}
\end{table}



\vspace{-1em}
\subsection{Domains}

We consider a discretized $d$-dimensional domain representing the full observable space
\[
\mathcal{C}_N^{(d)} = \{0,\dots,N\}^d,
\qquad |\mathcal{C}_N^{(d)}| = M = (N+1)^d.
\]

\noindent
Within this cube, we define the simplex-constrained inner domain
\[
\mathcal{I}_N^{(d)} =
\{x \in \mathcal{C}_N^{(d)} : \|x\|_1 \le N\},
\qquad
|\mathcal{I}_N^{(d)}| = K = \binom{N+d}{d}.
\]

\noindent
The remaining points form the outer domain
\[
\mathcal{O}_N^{(d)} =
\mathcal{C}_N^{(d)} \setminus \mathcal{I}_N^{(d)},
\qquad
|\mathcal{O}_N^{(d)}| = M-K.
\]

\noindent
The observable cube naturally decomposes into discrete shells indexed by the
$\ell_1$ norm
\[
s(x)=\|x\|_1.
\]
We denote by
\[
\pi_{\mathrm{shell}} : \mathcal{C}_N^{(d)}
\rightarrow \{1,\dots,M\}
\]
a shell permutation which orders configurations by increasing shell index,
while preserving a fixed internal ordering within each shell. This decomposition 
separates a structured combinatorial core from its surrounding observable domain.

\vspace{0.4em}

\noindent
We introduce the combinatorial configuration space
\[
\Phi = \{0,1,\dots,d-1\}^N,
\]
whose elements represent discrete sequences (paths) of length $N$ over $d$ symbols.
Each configuration $\phi \in \Phi$ induces prefix-count coordinates satisfying
\[
x(\phi) \in \mathcal{C}_N^{(d)},
\qquad
x_i(\phi)=\#\{k\le N:\phi_k=i\},
\]
with
\[
\|x(\phi)\|_1=N.
\]

\vspace{0.4em}

\noindent
The tensor $T$ aggregates contributions over configurations
$\phi \in \Phi$ mapping to identical observable coordinates, thereby inducing
a structured relation between simplex-constrained latent configurations and
their observable extensions.

\subsection{Tensor Construction}
\label{sec:tensor_construction}

We provide a geometric interpretation of the CPrefix tensor.
The transparent cube represents the observable domain whose coordinates
$(R,G,B)$ act as observable thresholds:
\[
\mathcal{C}_N^{(3)}=\{0,\ldots,N\}^3,
\qquad
|\mathcal{C}_N^{(3)}|=(N+1)^3.
\]

\noindent
This cube supports multiple levels of observation.
While its interior corresponds to full multi-channel observations,
the representation is not injective: each observable point aggregates
multiple underlying configurations compatible with the simplex constraint.
The cube boundary (faces, edges, and vertices) supports contracted observables
arising from marginalizations of the simplex-constrained inner domain.

\vspace{0.5em}
\noindent
The construction is driven by an underlying combinatorial process
defined on ternary paths
\[
\Phi=(\phi_1,\ldots,\phi_N)\in\{0,1,2\}^N,
\qquad
|\{0,1,2\}^N|=3^N,
\]
which can be interpreted as lattice paths on the discrete Pascal simplex.

\vspace{0.5em}
\noindent
For each path $\Phi$ and observable threshold $(R,G,B)$,
we define inner coordinates through prefix-count projections:
\[
\begin{aligned}
{\textcolor{redC}r} &= R_{\Phi}({\textcolor{redC}R}) = \#\{i\le {\textcolor{redC}R}:\phi_i=0\},\\
{\textcolor{greenC}g} &= G_{\Phi}({\textcolor{greenC}G}) = \#\{i\le {\textcolor{greenC}G}:\phi_i=1\},\\
{\textcolor{blueC}b} &= B_{\Phi}({\textcolor{blueC}B}) = \#\{i\le {\textcolor{blueC}B}:\phi_i=2\}.
\end{aligned}
\]

\noindent
The accumulated counts are constrained by the observable thresholds:
\vspace{-0.5em}
\[
\min(\textcolor{redC}{R},\textcolor{greenC}{G},\textcolor{blueC}{B})
\;\le\;
\textcolor{redC}{r}+\textcolor{greenC}{g}+\textcolor{blueC}{b}
\;\le\;
\max(\textcolor{redC}{R},\textcolor{greenC}{G},\textcolor{blueC}{B}).
\]

\noindent
The three inner coordinates correspond to symbol-wise counting projections
of the same path $\Phi$, yielding $(r,g,b)$ as a decomposition of its
accumulated counts. They are therefore not independent quantities,
but coupled observable projections of a single combinatorial process.

\vspace{0.5em}
\noindent
Applying this construction to all paths
$\Phi\in\{0,1,2\}^N$
induces, for each observable threshold $(R,G,B)$,
a distribution over admissible inner configurations $(r,g,b)$.
These configurations lie in the simplex-constrained inner domain
\[
\mathcal{I}_N^{(3)}=
\{(r,g,b)\in\mathcal{C}_N^{(3)}:
{\textcolor{redC}r}+{\textcolor{greenC}g}+{\textcolor{blueC}b}\le N\},
\]
with cardinality
\vspace{-0.5em}
\[
K = |\mathcal{I}_N^{(3)}|
= \binom{N+3}{3}.
\]
\noindent
We define the observable tensor
\[
T({\textcolor{redC}R},{\textcolor{greenC}G},{\textcolor{blueC}B},{\textcolor{redC}r},{\textcolor{greenC}g},{\textcolor{blueC}b}),
\]
as the multiplicity of ternary configurations compatible with a given observable state:
\vspace{-0.5em}
\begin{align*}
T({\textcolor{redC}R},{\textcolor{greenC}G},{\textcolor{blueC}B},{\textcolor{redC}r},{\textcolor{greenC}g},{\textcolor{blueC}b})
&=
\bigl|\{\Phi\in\{0,1,2\}^N : {} \\
&\quad R_\Phi({\textcolor{redC}R})={\textcolor{redC}r},\quad G_\Phi({\textcolor{greenC}G})={\textcolor{greenC}g},\quad B_\Phi({\textcolor{blueC}B})={\textcolor{blueC}b} \}\bigr|.
\end{align*}

\noindent
The tensor therefore acts as a structured observation operator linking
observable states in the cube
$\mathcal{C}_N^{(3)}$
to admissible configurations in the simplex
$\mathcal{I}_N^{(3)}$.
Each observable coordinate $(R,G,B)$ induces a distribution over the latent simplex, revealing the underlying combinatorial organization of the observable domain.

\vspace{0.5em}
\noindent
A shell-aggregated observable is obtained by contracting simplex layers:
\vspace{-0.5em}
\[
T(R,G,B,s)
=
\sum_{\substack{r,g,b\ge0\\ r+g+b=s}}
T(R,G,B,r,g,b),
\]
where
\(
s=r+g+b
\)
indexes diagonal shells of the trinomial Pascal simplex. A further contraction over observable shells yields
\[
T(S,s),
\qquad
S=R+G+B,
\]
leading to the hierarchy
\[
\boxed{
T(R,G,B,r,g,b)
\;\rightarrow\;
T(R,G,B,s)
\;\rightarrow\;
T(S,s)}
\]

\subsection{Normalization}

\noindent
Since the tensor is induced by the uniform ensemble
\[
\Phi\in\{0,1,2\}^N,
\qquad
|\{0,1,2\}^N|=3^N,
\]
its entries satisfy
\[
\sum_{r,g,b}
T(R,G,B,r,g,b)
=
3^N.
\]

\noindent
The associated normalized tensor is therefore
\[
T_{\mathrm{prob}}(R,G,B,r,g,b)
=
\frac{
T(R,G,B,r,g,b)
}{
3^N
},
\]
which defines a probability distribution over the simplex domain
$\mathcal{I}_N^{(3)}$.

\noindent
Normalization is preserved under tensor contractions, yielding the induced distributions
\[
\boxed{
T_{\mathrm{prob}}(R,G,B,r,g,b)
\;\rightarrow\;
T_{\mathrm{prob}}(R,G,B,s)
\;\rightarrow\;
T_{\mathrm{prob}}(S,s)}
\]
\vspace{-0.5em}

\section{Observable Geometry and Reconstruction}
The combinatorial tensor induces a structured linear operator
between the simplex-constrained latent domain and the observable cube.
Unlike classical multilinear tensor frameworks~\cite{KoldaBader2009},
its entries arise from combinatorial observability constraints.
After shell ordering by $\pi_{\mathrm{shell}}$, the flattened tensor admits the block decomposition
\[
T=
\begin{pmatrix}
T_{\mathrm{II}}\\
T_{\mathrm{OI}}
\end{pmatrix},
\]
where \(T_{\mathrm{II}}\) describes interactions internal to the admissible
simplex and \(T_{\mathrm{OI}}\) its extension into the outer observable
domain. The induced observable-to-latent geometry and the distinction between
inner and outer shell mappings are visualized in
\textbf{Fig.~\ref{fig:ObservableLatentTensor}}.

\vspace{0.25em}
\noindent
Given latent simplex coefficients
\[
C_{\mathrm{inner}}\in\mathbb{R}^{K},
\]
the observable representation is
\[
X_{\mathrm{obs}}
=
T\,C_{\mathrm{inner}},
\qquad
X_{\mathrm{obs}}\in\mathbb{R}^{M}.
\]

\vspace{0.25em}
\noindent
The associated kernel~\cite{ShaweTaylor2004}
\[
K_T=T^\top T
\]
induces a geometry in which latent configurations are compared through
their observable projections. The tensor therefore acts as a structured
observable operator linking admissible latent configurations to measurable
states.

\vspace{0.25em}
\noindent
The induced block structure exhibits a strong asymmetry between
the admissible simplex and its observable lift. Numerically,
\[
\operatorname{rank}(T)=K,
\qquad
\operatorname{rank}(T_{\mathrm{II}})<K,
\qquad
\operatorname{rank}(T_{\mathrm{OI}})=K.
\]
Thus, the simplex core alone is not fully self-identifying,
whereas the outer observable region provides a full-rank encoding
of admissible configurations. Full identifiability therefore
emerges through the outer lift.

\begin{figure}[H]
\centering
\includegraphics[width=0.8\columnwidth]{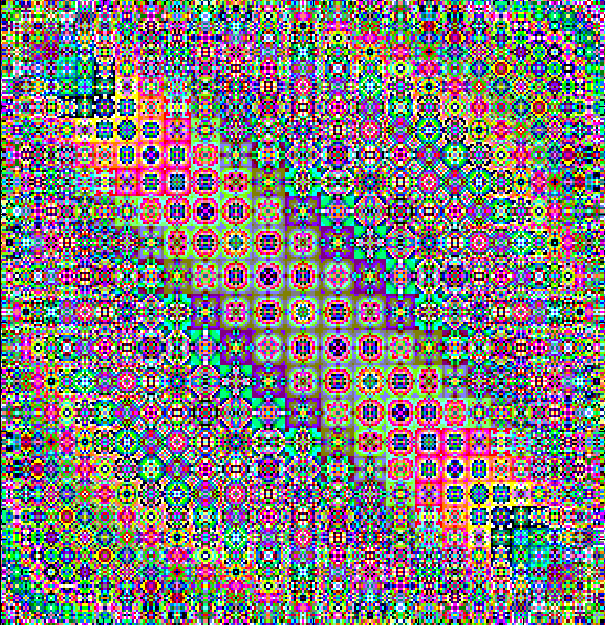}
\caption{
\footnotesize{
\textbf{Geometry induced by the observable representation.}}
Visualization of
$\Re(\log(T\,\mathrm{pinv}(T)))$
after shell ordering. The projection operator captures how observable
configurations overlap through their common latent combinatorial
representation, revealing recursive shell organization and spectral-like
combinatorial structures on the Pascal simplex. Colors are derived from 
a uniform sampling of the RGB color space and provide a perceptual 
rendering of the shell-ordered projection without encoding information beyond the underlying combinatorial geometry.
}
\label{fig:kernelLog}
\vspace{-1em}
\end{figure}

\vspace{0.25em}
\noindent
The observable representation induces a geometry through the projection operator
\[
P_T=T\,\mathrm{pinv}(T).
\]
The projection captures how observable configurations overlap through
their common latent combinatorial representation.
\textbf{Fig.~\ref{fig:kernelLog}} visualizes
\(
\Re\!\bigl(\log(P_T)\bigr),
\)
whose logarithmic form reveals the intrinsic combinatorial geometry
induced by the observable representation, exposing recursive shell
organization and spectral-like coupling structures on the Pascal simplex.

\subsection{Latent Reconstruction}

Observable measurements can be projected back onto the admissible simplex
through a regularized inverse problem~\cite{Hansen2010}:
\[
C_{\mathrm{inner}}
=
(T^\top T+\lambda I)^{-1}T^\top X_{\mathrm{obs}},
\]
with \(\lambda>0\) ensuring numerical stability.

The reconstructed observable estimate is then
\[
X_{\mathrm{recon}}
=
T\,C_{\mathrm{inner}}.
\]

Unlike generic linear inversion, the reconstruction remains constrained by
the intrinsic simplex geometry and shell organization induced by the tensor
construction.

\subsection{Color Interpretation}

Color imaging provides a representative application for validating the proposed observable representation. Rather than benchmarking existing color-management pipelines, the objective is to evaluate whether measured color transformations admit faithful combinatorial observable representations. Reconstruction accuracy therefore reflects both numerical fidelity and the compatibility of the underlying mapping with the proposed framework.

\vspace{0.25em}
\subsubsection{ICC-Based Color Reconstruction}

ICC profiles provide high-quality measured mappings from uniformly sampled RGB configurations to CIE XYZ values. Uniform RGB samples define the observable domain \(\mathcal{C}_N^{(3)}\), while admissible configurations satisfying \(r+g+b\le N\) define the latent combinatorial representation \(\mathcal{I}_N^{(3)}\). The measured XYZ responses constitute the observable mapping, from which latent coefficients are recovered through the inverse tensor formulation.

\vspace{0.25em}
\noindent
Experiments were conducted on display and photographic printer ICC profiles from Apple, Epson, HP, and Canon. Printer profiles were provided courtesy of Hahnemühle Photo Rag®~\cite{hahnemuhle_icc,ICC}. Although driven by RGB inputs, the printer systems internally operate on proprietary multi-ink device spaces with nonlinear channel interactions.

\vspace{0.25em}
\noindent
Reconstruction quality was evaluated using channel-wise XYZ RMSE together with perceptual differences in CIE~Lab space through \(\Delta E_{00}\)~\cite{Sharma2005CIEDE2000} (mean / 95\% / max). Experiments were performed on RGB grids of size \(M=4913,\ K=969\) for \(N=16\), and \(M=39304,\ K=7140\) for \(N=33\).

\captionsetup[table]{labelformat=empty}
\begin{table}[!h]
\vspace{-0.5em}
\centering
\caption[]{
\footnotesize{Tensor reconstruction accuracy for display and printer ICC profiles.}
}
\scriptsize
\setlength{\tabcolsep}{3pt}
\renewcommand{\arraystretch}{1.12}

\begin{tabular}{llccc}
\toprule
Profile & Domain & $N$ & RMSE$_{XYZ}$ & $\Delta E_{00}$ \\
\midrule

Apple Display P3 & Display RGB & 16
& [9e-06 1e-05 1e-05]
& .001 / .002 / .005 \\

\midrule

PSO Coated v3 & CMYK Offset & 16
& [.0006 .0007 .0005]
& .155 / .351 / .848 \\

\midrule

EPSON P9000 & RGB Printer & 16
& [.00188 .0020 .0019]
& .269 / .644 / 2.556 \\

EPSON P9000 & RGB Printer & 33
& [.0009 .0009 .0009]
& .116 / .252 / 1.923 \\

HP Z3200 & RGB Printer & 16
& [.0043 .0045 .0043]
& .523 / 1.198 / 4.053 \\

HP Z3200 & RGB Printer & 33
& [.0021 .0023 .0021]
& .217 / .509 / 2.958 \\

CANON Pro4100 & RGB Printer & 16
& [.0017 .0017 .0020]
& .234 / .512 / 2.505 \\

CANON Pro4100 & RGB Printer & 33
& [.0008 .0008 .0010]
& .095 / .218 / 1.374 \\

\bottomrule
\end{tabular}

\label{tab:icc_tensor_results}
\vspace{-0.5em}
\end{table}

\noindent
The results demonstrate that measured color transformations admit accurate observable representations within the proposed framework. The \(\Delta E_{00}\) statistics quantify the perceptual deviation from the original ICC transformations, with most reconstructed colors remaining below the commonly accepted visibility threshold (\(\Delta E_{00}<1\)), indicating near-imperceptible deviations across most of the sampled device gamut.

\vspace{0.25em}
\noindent
The display profile exhibits near-perfect reconstruction, reflecting the
comparatively smooth and low-dimensional structure of RGB display
transforms.

\vspace{0.25em}
\noindent
In contrast, the printer profiles involve hidden multi-ink
interactions and stronger nonlinearities, yet remain accurately represented within the proposed observable representation.

\vspace{0.25em}
\noindent
The PSO Coated v3 profile exhibits lower reconstruction errors than the photographic printer profiles, reflecting the greater structural regularity of standardized FOGRA/PSO offset-printing characterization data.

\vspace{0.25em}
\subsubsection{Perceptual Gamut Mapping}
Building upon the previous ICC reconstruction results, the tensor framework
naturally induces a structured perceptual gamut mapping~\cite{Morovic2008}  between latent
simplex representations. Given two device profiles, latent coefficients
\[
C_A,\; C_B \in \mathbb{R}^{K\times3}
\]
are first recovered from their observable XYZ responses through the inverse
tensor formulation. A structured transport operator is then estimated
directly in the latent simplex domain,
\[
C_B \approx M\,C_A,
\]
yielding a perceptual mapping between admissible configurations rather than
a direct interpolation in observable color space.

\vspace{0.25em}
\noindent
Preliminary experiments investigated several latent transport models. Current
results were obtained using localized shell-band transport with limited
cross-shell interactions (\(\Delta s=\pm2\)), suggesting that perceptual
gamut compression remains predominantly shell-local within the latent
simplex geometry.

\vspace{0.25em}
\noindent
For a Display~P3 to EPSON~P9000 / Hahnemühle Photo Rag® printer-profile
mapping at \(N=33\), the proposed shell-band model achieved for mean / \(95^{\mathrm{th}}\) percentile / maximum error respectively.
\[
\Delta E_{00}
=
0.38 \; / \; 0.82 \; / \; 3.22
\]

\begin{figure}[H]
\vspace{-0.75em}
\centering
\includegraphics[width=0.8\columnwidth]{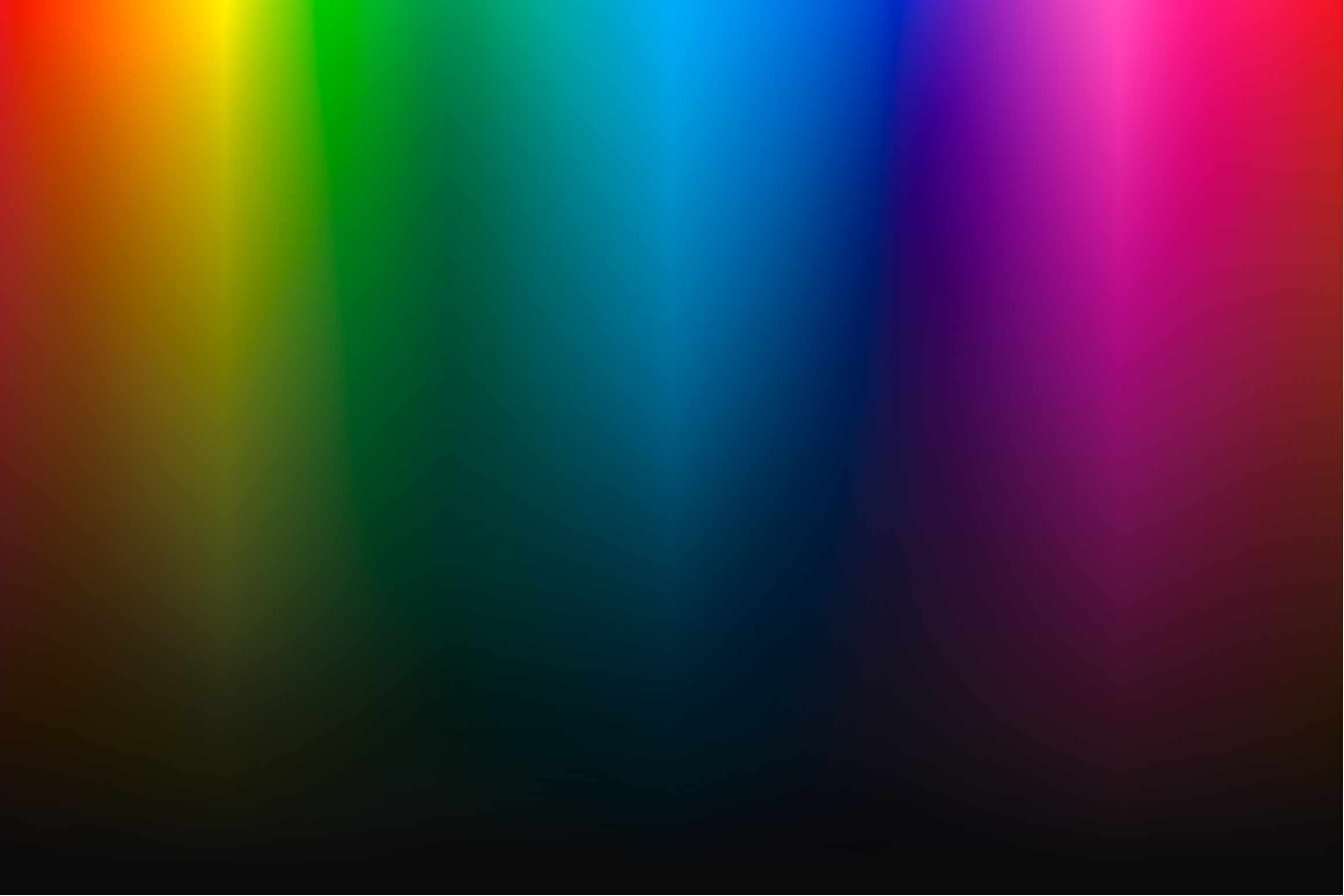}
\caption{
\footnotesize{
\textbf{CPrefix perceptual gamut mapping on Granger chart.}
The result illustrates the smooth shell-local compression induced by the latent simplex transport.
Current transport behavior favors global continuity and shell smoothness
over strict local chroma preservation.
}}
\label{fig:Granger}
\vspace{-0.75em}
\end{figure}


\begin{figure*}[b]
\centering
\includegraphics[width=0.70\textwidth]{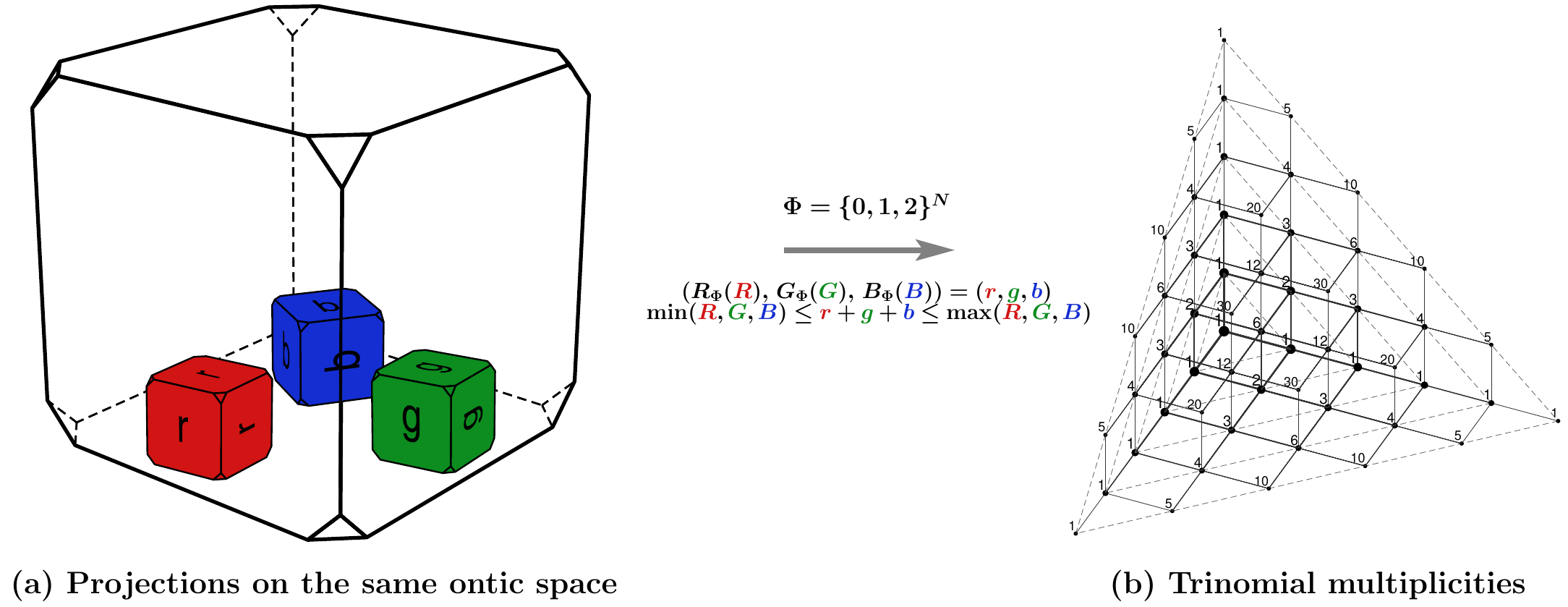}
\caption{
\footnotesize{
\textbf{Combinatorial projections and trinomial organization on a ternary observable space.}
(a) Three observable components $(R,G,B)$ act on the common discrete space
$\Phi=\{0,1,2\}^N$, generating cumulative observable counts projected onto
shared lattice coordinates.
(b) The same discrete space induces admissible trinomial configurations $(r,g,b)$ whose multiplicities follow a Pascal-simplex geometry, forming the support of a multinomial distribution. This reveals how a uniform ensemble of discrete trajectories gives rise to a structured combinatorial distribution that is lifted into the observable multi-channel domain through the tensor mapping.
}}
\label{fig:CQDICE}
\vspace{1.0em}
\centering
\includegraphics[width=0.70\textwidth]{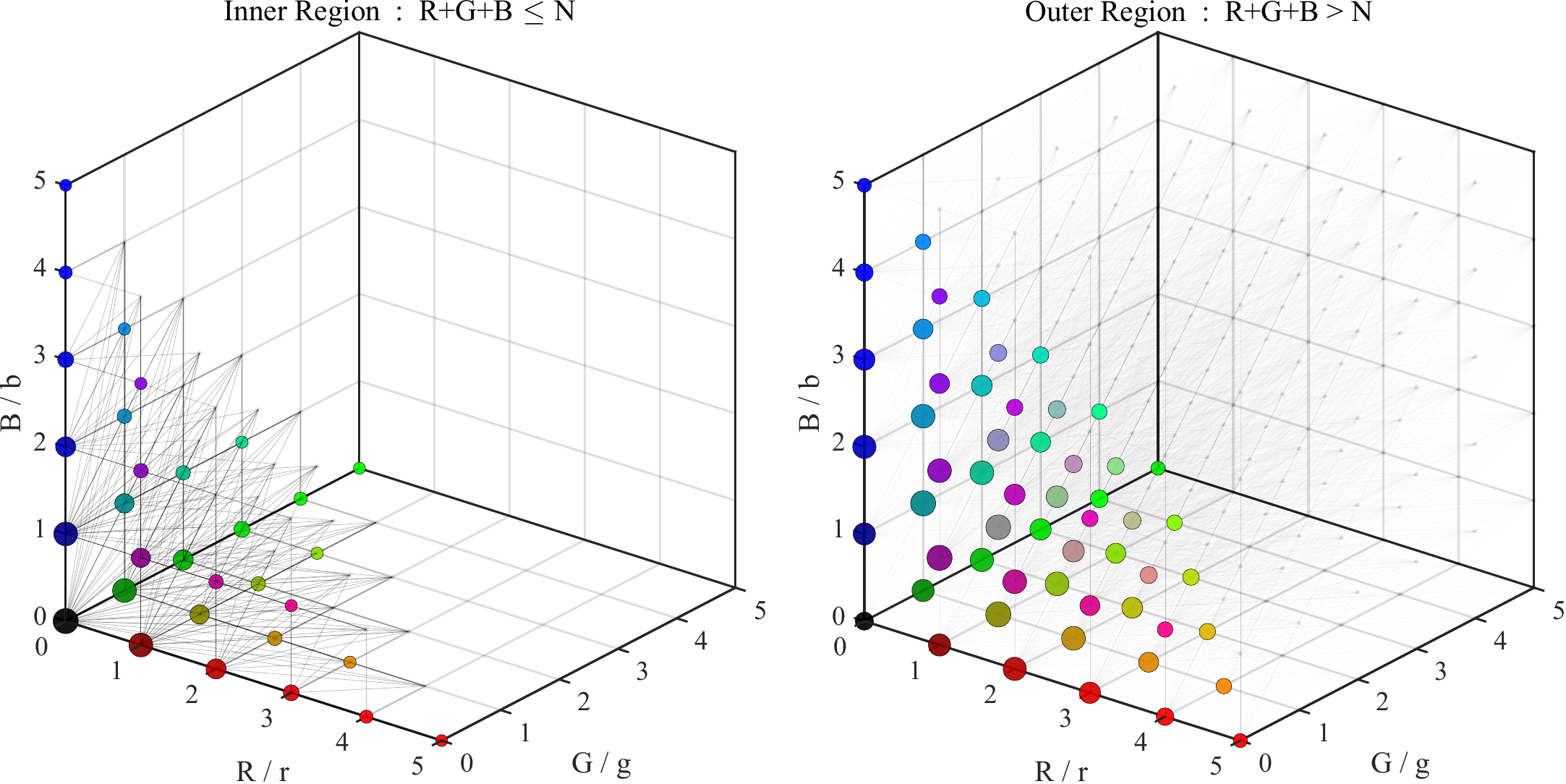}
\caption{
\footnotesize{
\textbf{Observable-to-Latent Tensor Geometry --- Inner and Outer Shell Mappings for $N=5$.}
The figure visualizes the structured mapping induced by the CPrefix tensor between the observable cube
$\mathcal{C}_N^{(3)}$ and the simplex-constrained latent domain
$\mathcal{I}_N^{(3)}$.
The $(R,G,B)$ represent observable cube coordinates, whereas colored
spheres correspond to latent simplex coordinates $(r,g,b)$, with sphere colors
directly induced from normalized latent RGB coordinates and sphere sizes
proportional to the accumulated tensor mass. The left panel shows mappings associated with the admissible inner region
defined by $R+G+B \leq N$, while the right panel displays mappings originating
from the outer observable shell $R+G+B > N$.
Tensor edges visualize the combinatorial transport between observable and
latent configurations, revealing the geometric organization of the inner
simplex core and its observable outer extension.
}}
\label{fig:ObservableLatentTensor}
\end{figure*}

\section{Conclusion}
CPrefix introduces a combinatorial observable representation for structured discrete mappings. By lifting a discrete Pascal simplex into an observable domain, the framework separates the combinatorial organization of a mapping from its measured values, providing a unified representation in which observable behavior can be represented, reconstructed, and analyzed.

\vspace{0.25em}
\noindent
The tensor formulation establishes a structured relationship between latent combinatorial configurations and observable measurements, enabling faithful reconstruction together with a quantitative characterization of the underlying mapping. 

\vspace{0.25em}
\noindent
Experiments on practical ICC workflows demonstrate that measured color transformations admit accurate observable representations while naturally supporting structured perceptual gamut transport. Reconstruction residuals further provide insight into the compatibility of measured mappings with the proposed representation.

\vspace{0.25em}
\noindent
Although demonstrated on color imaging, the construction is independent of the physical interpretation of the observables and naturally extends to arbitrary structured multi-channel mappings. Potential applications include spectral reconstruction, multi-ink printer separation, metameric analysis, structured inverse problems, and other domains where observable measurements arise from constrained combinatorial configurations.

\vspace{0.25em}
\noindent
More generally, the proposed framework suggests that combinatorial observable representations constitute a useful mathematical language for reasoning about structured discrete mappings. Future work will investigate higher-dimensional observable systems, generalized observable representations, perceptual transport operators, and broader applications across computational imaging and structured sensing.
\clearpage

\bibliographystyle{IEEEtran}
\bibliography{IEEEabrv,references}

\clearpage
\onecolumn
\appendix

\section{Additional Tensor Folding Visualization}

The following figure illustrates the complete sequence of tensor transformations used to derive the compressed observable representation described in Section~\ref{sec:tensor_construction}. It complements Figs.~\ref{fig:ObservableLatentTensor} by displaying the intermediate folding operations that transform the refined tensor $T(R,G,B,r,g,b)$ into the shell representation $T(S,s)$.

\begin{figure*}[!h]
    \vspace{0.0em}
    \centering
    \includegraphics[width=1\textwidth]{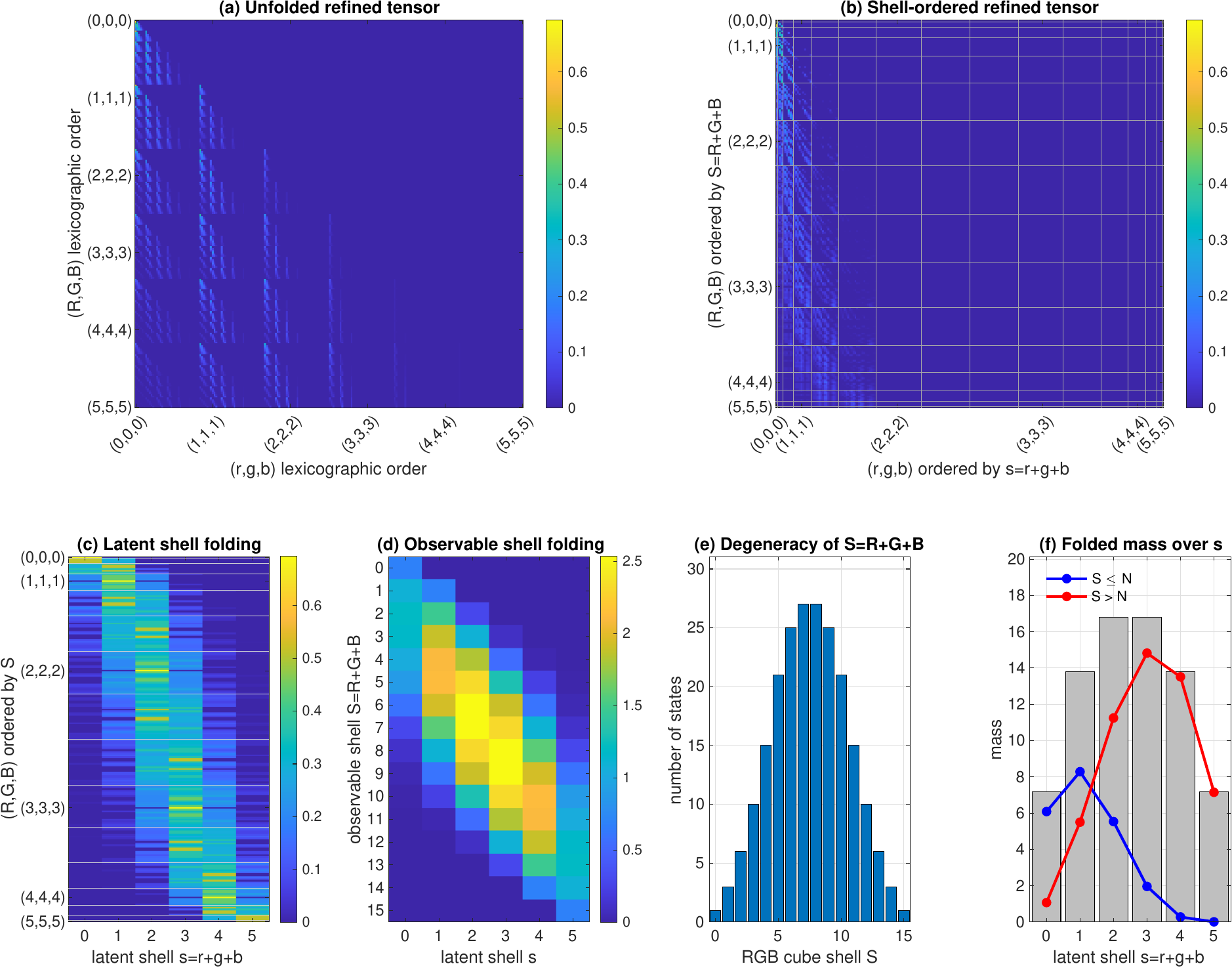}
\caption{
\textbf{Observable folding hierarchy of the refined RGB tensor construction for $N=5$.}
(a) Unfolded refined tensor $T(R,G,B,r,g,b)$ displayed in lexicographic ordering of both observable RGB states $(R,G,B)$ and latent multinomial states $(r,g,b)$. The tensor exhibits sparse admissibility structures induced by the latent combinatorial organization associated with the discrete RGB lattice.
(b) Shell-ordered tensor obtained by reindexing both domains according to the shell variables $S=R+G+B$ and $s=r+g+b$. The permutation reveals an emergent block geometry organized by shell degeneracies and admissible observable transitions. White separator lines indicate shell boundaries.
(c) Latent shell folding $(r,g,b)\rightarrow s=r+g+b$, aggregating latent multinomial configurations into trinomial shell coordinates while preserving the observable RGB ordering. The resulting representation exposes structured aggregation patterns induced by latent multinomial shell organization.
(d) Observable shell folding $(R,G,B)\rightarrow S=R+G+B$, yielding the compressed operator $T(S,s)$. The resulting matrix reveals a Pascal-like shell geometry coupling observable RGB shells to latent multinomial shells.
(e) Degeneracy of the observable RGB shells as a function of $S=R+G+B$. The histogram exhibits the characteristic triangular profile of the multinomial shell degeneracies.
(f) Total folded tensor mass as a function of the latent shell index $s$. Gray bars represent the total shell mass obtained after folding, while the blue and red curves separate the contributions arising from observable shells satisfying $S\leq N$ and $S>N$, respectively. The plot highlights how the folded representation redistributes observable contributions across latent multinomial shells.
}
\label{fig:tensor_full_folding}
\end{figure*}

\clearpage

The following figure complements Fig.~\ref{fig:tensor_full_folding} by comparing the latent-shell tensor with a photometric observable defined on the same shell-ordered RGB configuration space. Despite their different constructions, the two representations exhibit a remarkably similar global geometry. The latent-shell tensor arises from the symmetric aggregation $s=r+g+b$, whereas the photometric projection assigns to each observable RGB state $(R,G,B)$ the weighted scalar value $Y=\operatorname{round}(0.299R+0.587G+0.114B)$. This structural similarity suggests a natural extension of the combinatorial construction in which ternary paths are endowed with channel-dependent weights, potentially providing a path-based interpretation of such photometric projections.
\begin{figure*}[!h]
    \vspace{1.0em}
    \centering
    \includegraphics[width=1\textwidth]{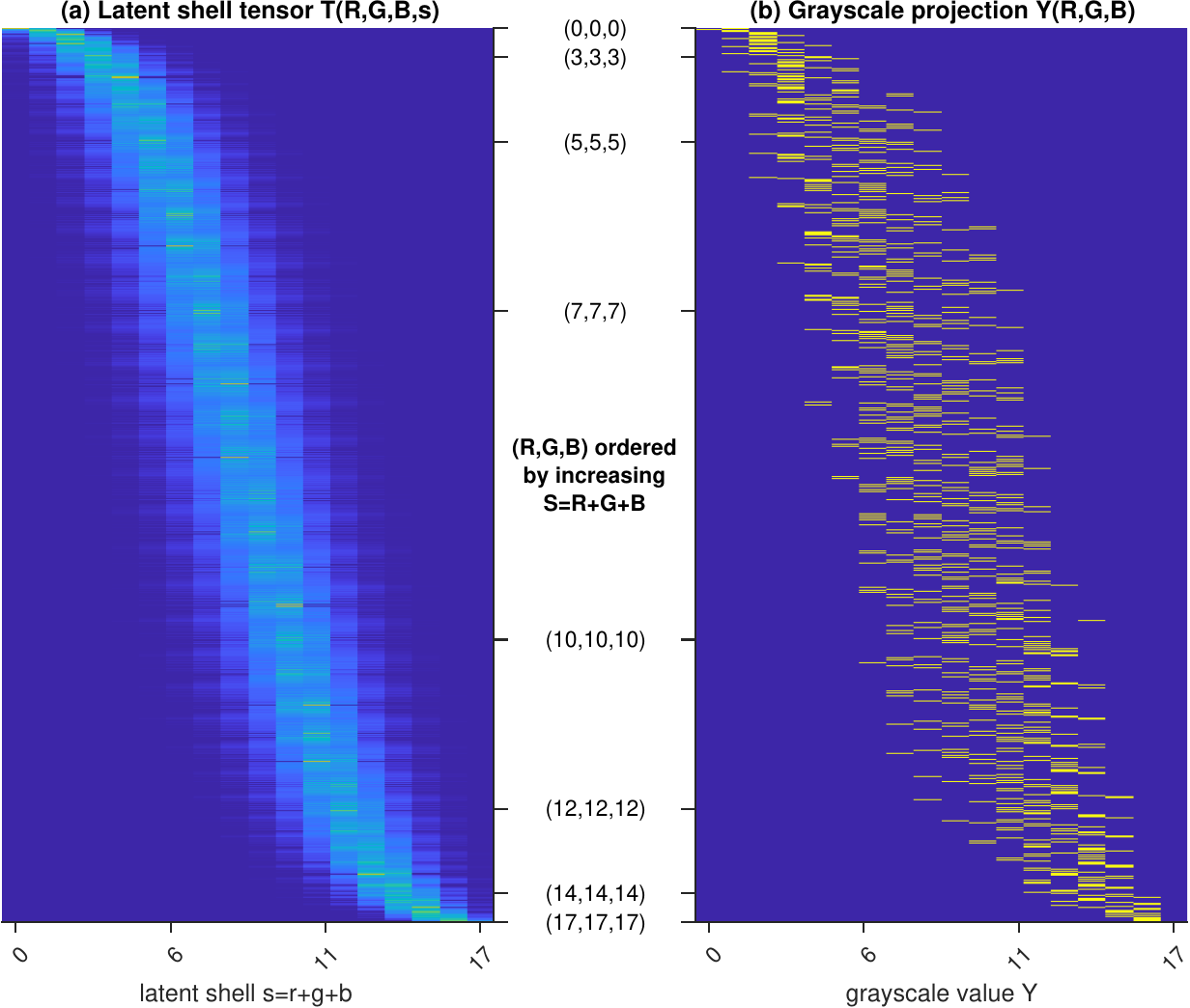}
\caption{
\textbf{Latent-shell tensor and photometric observable projection of the same RGB configuration space for $N=17$.}
(a) Folded tensor $T(R,G,B,s)$, obtained by aggregating latent configurations $(r,g,b)$ according to the shell coordinate $s=r+g+b$. Observable RGB configurations $(R,G,B)$ are ordered by increasing shell index $S=R+G+B$, with selected achromatic configurations shown as reference points.
(b) Projection of the same shell-ordered RGB configurations onto a scalar photometric observable. Grayscale values are computed as $Y=\operatorname{round}(0.299R+0.587G+0.114B)$, following Rec.~ITU-R BT.601-7. The two panels therefore expose the combinatorial and photometric organization of the same ordered configuration space.
}
\label{fig:latent_shell_photometric}
\end{figure*}

\end{document}